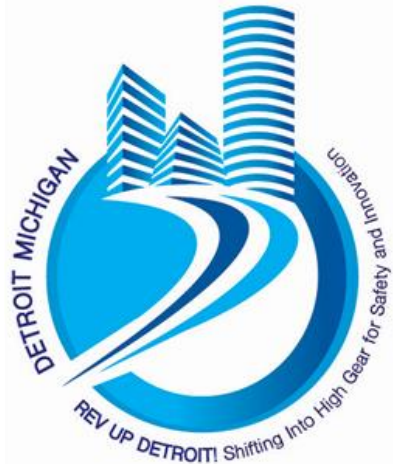

**OBSTACLE DETECTION TESTS IN REAL-WORLD TRAFFIC CONTEXTS FOR THE PURPOSES OF MOTORCYCLE AUTONOMOUS EMERGENCY BRAKING (MAEB)**

**Giovanni Savino**
**Simone Piantini**
Gustavo Gil
**Marco Pierini**
University of Florence
Italy

**ABSTRACT**

Research suggests that a Motorcycle Autonomous Emergency Braking system (MAEB) could influence 25% of the crashes involving powered two wheelers (PTWs). By automatically slowing down a host PTW of up to 10 km/h in inevitable collision scenarios, MAEB could potentially mitigate the crash severity for the riders. The feasibility of automatic decelerations of motorcycles was shown via field trials in controlled environment. However, the feasibility of correct MAEB triggering in the real traffic context is still unclear. In particular, MAEB requires an accurate obstacle detection, the feasibility of which from a single track vehicle has not been confirmed yet. To address this issue, our study presents obstacle detection tests in a real-world MAEB-sensitive crash scenario.

## INTRODUCTION

When talking about Autonomous Emergency Braking applied to powered two wheelers (PTWs), one common issue raised both by researchers and users is the practicability of an abrupt deceleration deployed by the system without inputs from the rider. From a technical point of view though, in the light of current ABS systems, applying an automatic braking appears straightforward; the critical element is to perform a reliable obstacle detection from the single-track vehicle due to its physiological tilting. In this paper, we will present our findings regarding a test of obstacle detection in the real traffic while emulating the pre-crash phase of a real-world crash case.

### Background on MAEB

A motorcycle AEB (MAEB) is a system that detects inevitable collision scenarios and deploys an automatic braking manoeuvre of the motorcycle (or more in general, the PTW) also without a direct braking input from the rider. The speed reduction at impact produced by MAEB could potentially mitigate the crash severity for the riders.

According to previous studies, MAEB could influence approximately one fourth of the crashes involving PTWs [1]. The analysis of the effects of MAEB was conducted with 2D computer simulations of sets of real world crashes [2, 3]. These simulations showed that when assuming a conservative approach for the activation (namely, triggering after the collision becomes inevitable and limiting the target automatic deceleration to 0.3 $g$ when the rider does not apply any braking), the typical effect of MAEB is to reduce the impact speed of the motorcycle by 4 km/h (see Figure 1). In some cases, the theoretical impact speed reduction was up to 10 km/h. The authors also evaluated MAEB effects assuming: i) an ideal obstacle detection system; and ii) a more realistic system with limited field of view in terms of angle and range. The latter configuration was not found to limit MAEB influence except for a few cases [3]. An explanation derives from the criterion of inevitable collision state used for the triggering, which limits the system to intervene less than 0.4 s before the actual collision. So at that point in time that the obstacle has already entered the field of view.

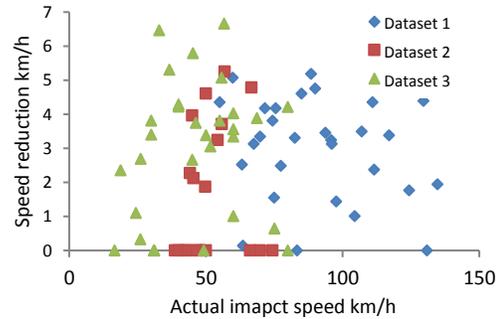

*Figure 1. Effects of MAEB in terms of estimated speed reduction vs. original speed at impact for the in-depth crash cases from three datasets [2].*

### Objective

To the authors' best knowledge, one missing component in the development of MAEB is a confirmation of the actual possibility of an accurate obstacle detection performed in the real world with sensors mounted on single track vehicles, which are characterized by non-negligible roll angles also when travelling along straight segments of road. We questioned whether current technologies enable obstacle detection with sufficient detail for the purposes of MAEB triggering in real world crash situations. To contribute on that, our study presents obstacle detection tests in a MAEB-sensitive crash scenario.

## METHODS

The obstacle detection systems analyzed in this study were an automotive LIDAR (reference system) and three sets of stereo cameras. We conducted several field experiments which progressively increased the level of realism up to involve data collection in real traffic. In the latter test, we emulated the pre-crash phase of a real world multivehicle crash involving a PTW (at the exact road location were the crash occurred). Finally, our results were compared against a 3D simulation experiment aiming to evaluate the performance of the imaging systems.

### Selection of the case study

First, a suitable set of case studies were identified from a subset of cases extracted from InSAFE, the in-depth crash investigation database active in the area of Florence, Italy [4]. The criteria for a case to be considered in the study were the following: i) PTW colliding against another vehicle (car or van);

ii) a 2D reconstruction of the vehicles' trajectories was available; iii) according to the results of a previous study [2], MAEB would have applied. For each selected case, details of MAEB activation were available from the cited study, including the time to collision (TTC) at which MAEB would have triggered, the reduction of speed at the impact produced by MAEB, and the position of the host PTW and opponent vehicle's positions at the time of MAEB triggering.

**Drive-through test protocol**
The second step was to perform drive-through tests in the exact crash locations with an instrumented PTW and a test car for each selected case. In particular, we logged obstacle detection devices scanning the environment to check their capability to properly detect the heading of the opponent car for the purposes of MAEB. The vehicles were driven by the research investigators along the same trajectories that led to the collision. For the opponent vehicle speed, we adopted the same velocity as that in the crash case; a safety upper limit was set, based on the location and on the specific manoeuvre. The PTW speed was set lower than that in the real case to avoid an actual collision. Speed profiles were defined case by case to let the opponent vehicle move safely in front of the PTW. The PTW was maintained stationary when the case reconstruction was considered dangerous with a moving PTW (eg. stationary PTW at traffic light instead of moving PTW). Given the different speed profiles of the vehicles compared to the actual cases, the synchronization of the trajectories was done referring to the vehicles' locations at MAEB triggering. These points were identified with computer simulations and marked on the spot for the drive through. Reconstructions in real traffic were attempted only for the cases in which actual vehicle trajectories were safe to be repeated and did not require any breach of road rules. For example, U-turn drive through tests were not conducted in the actual location if such manoeuvre was not allowed at the crash site. For some of the crash cases, surrogate tests were conducted in a parking lot for safety reasons. All the tests were conducted in daylight and good weather conditions and in dry asphalt. This study was approved from the Human Research Ethics Committee of the University of Florence.

**Equipment**
The test PTW was an instrumented scooter (Malaguti Spidermax 500) equipped with inertial measurement unit (X-Sens), lidar (IBEO Lux), and a tailored imaging system consisting of six low-cost action cameras (Camkong). Both PTW and opponent car were equipped with DGPS units (GeoMax Zenith 20) for accurate position measurements. The imaging system is depicted in figure 2 (technical characteristics are given in Table 1).

The imaging system is a rectilinear six camera rig which forms a trifocal stereo vision system. Optical and mechanical considerations for the fixation to the motorcycle frame to guarantee a proper performance of the imaging system were inspired from the agricultural field in with intelligent tractors deal with vibrations due to the irregularities of the terrain [5, 6]. The longer baseline (between cameras I and VI) is used to detect obstacles in a far range close to the PTW traveling axis. Cameras II and V measure the heading of frontal obstacles in a middle range with a wider region than the previous pair of stereo cameras. The central cameras III and IV are used to measure the obstacle's heading in the near field with a wider angle. A modification of the triggering switch of each camera allowed a hardware synchronization of the video footage (Figure 4). Videos were recorded in 1920x1080 aspect ratio at 30 fps for post-processing purposes. Finally, a verification of the synchronization of the six videos was done using a simple clapboard and subsequent offline check with the Open Source Kinovea software.

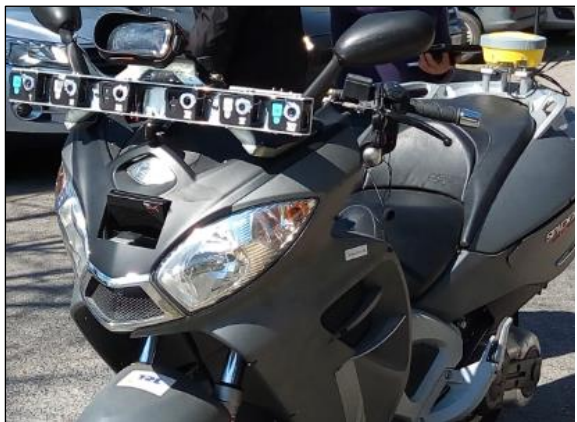

*Figure 2. Detail of the six cameras stereo rig anchored to the motorcycle frame by an inverted V-shape steel support. The stereo rig was placed 20 cm over the laser scanner, which is vertically aligned with the front wheel axle.*

**Table 1. Optical features of the trifocal stereo vision system.**

| Stereo pair | Baseline distance | Horizontal Field Of View | Focal length |
|---|---|---|---|
| I & VI | 597 mm | 80 deg | 1600 pix |
| II & V | 387 mm | 110 deg | 850 pix |
| III & IV | 149 mm | 170 deg | 950 pix |

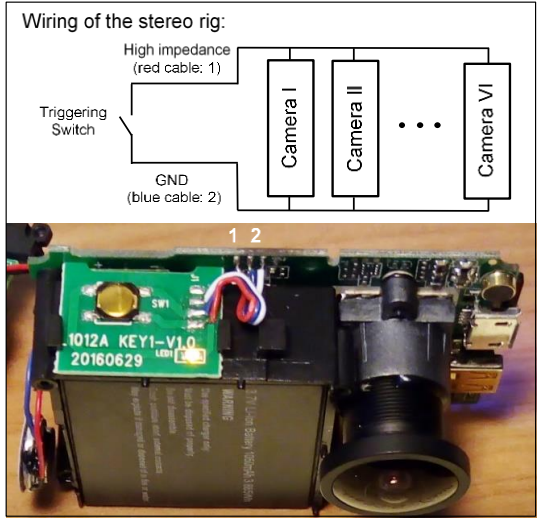

*Figure 4. Wiring detail of the six cameras. On top: the circuital scheme. Below: a picture of a disassembled camera showing the location of the electrical connections.*

The aim of the experiment was to observe whether MAEB could have properly detected the opponent vehicle for the purposes of triggering the emergency braking.

Our target was to measure the heading angle of the opposing vehicle in a real crash scenario occurred at an intersection, which requires peripheral perception. All the results presented in this paper will refer to the short range baseline (cameras III and IV). We will focus on the quality of the computed disparity maps because these are key to enable trajectory prediction of opposite vehicles without obstacle classification [7-11]. 3D point cloud reconstructions were generated from the information contained in the disparity maps and used as a measure to evaluate the quality of the disparity maps themselves obtained from our system.

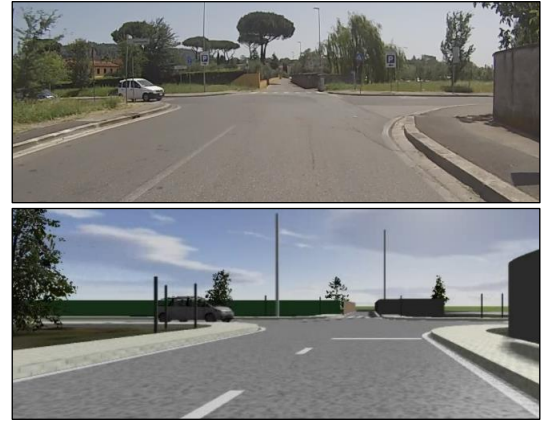

*Figure 3. Comparison between the real environment and the PreScan environment in a crash location.*

**3D spatial measurement**
Aiming to quantify the level of accuracy of the depth measurement system in daily light conditions, we reserved part of our office green area to build a calibration scene with 3D landmark that can be used as a referencial ground truth. The landmarks showed in Figure 6 were carrefully measured with a laser range finder (Leica Disto D5) and measuring tape.

**Computer simulations**
The crash cases were also recreated in a virtual environment using the software PreScan (TASS International). The road network at the crash locations was reconstructed in terms of road geometry and obstructions (including pavement, buildings, poles, traffic signs, walls, trees) to mimic the actual crash environment Figure 3. The trajectories and speed profiles of the host motorcycles and opponent vehicles were reproduced according to the original InSAFE crash reports. This computer environment allowed obtaining a synthetic 3D ground truth image similar to the one of the real scene. The virtual environment allowed to test the sensing methods with different speed profiles including those which led to the actual collision.

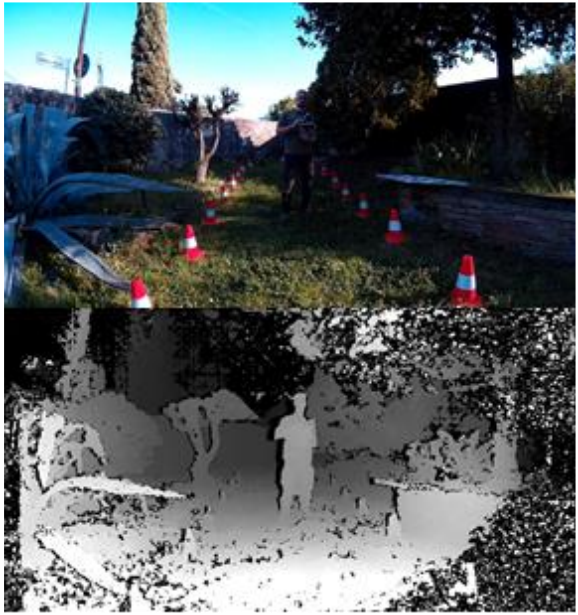

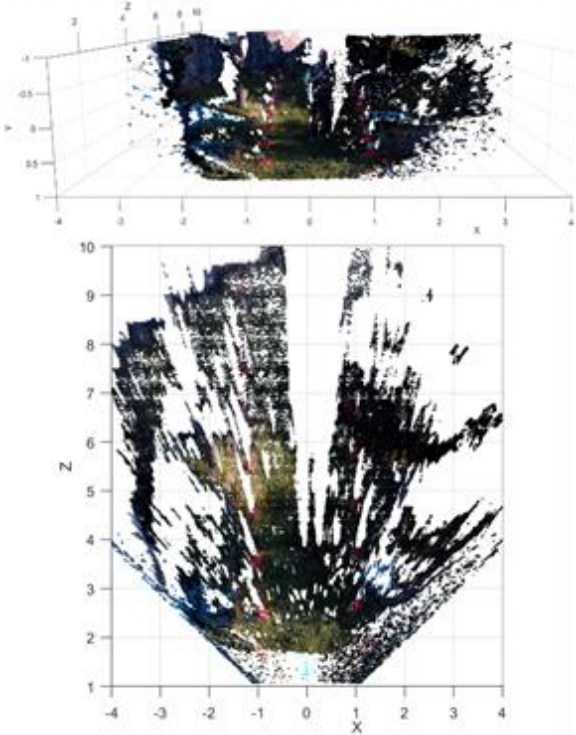

*Figure 5. Static calibration scene. The measurement of this reference scene was performed in different light conditions.*

**RESULTS**

The results of the stereo vision system will be presented in sets of 4 images. From top to bottom: 1) a rectified view taken from the left camera; 2) in gray scale the result of the dense disparity map calculation; 3) the 3D point cloud reconstruction of the scene; and 4) the top-view of the previous 3D point cloud showing the depth information of the scene.

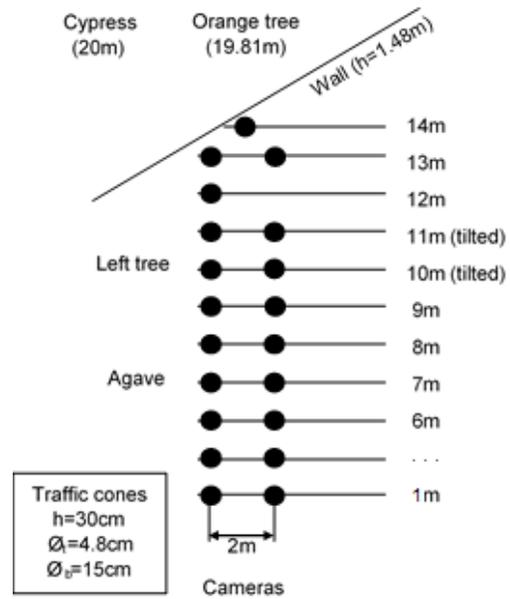

*Figure 6. Top view scheme of the location of the landmarks.*

The black color in the disparity map express no or unreliable disparity, meaning that this part of the scene is out of the measurement range of the imaging system or that the texture cannot be distinguished by the Semi-Global Matching (SGM) algorithm. Light gray colors express a large disparity in the stereo pair, meaning that this part of the scene is near to the imaging system.

Figure 5 illustrates the problem to properly determine the depth in top and bottom right corners. The non-uniform and abrupt transitions between black and white (noise) is an undesirable effect of depth ambiguities due to the similar texture of this zone. Specialized literature investigated this effects long time ago [6] and algorithms for urban scenes were developed [7].

**Refining the heading measurement**
To obtain the heading angle of the opposite vehicles in real traffic situations, we selected a small vehicle (Fiat Panda) assuming that smaller vehicles represent the worst case scenario for remote sensing.

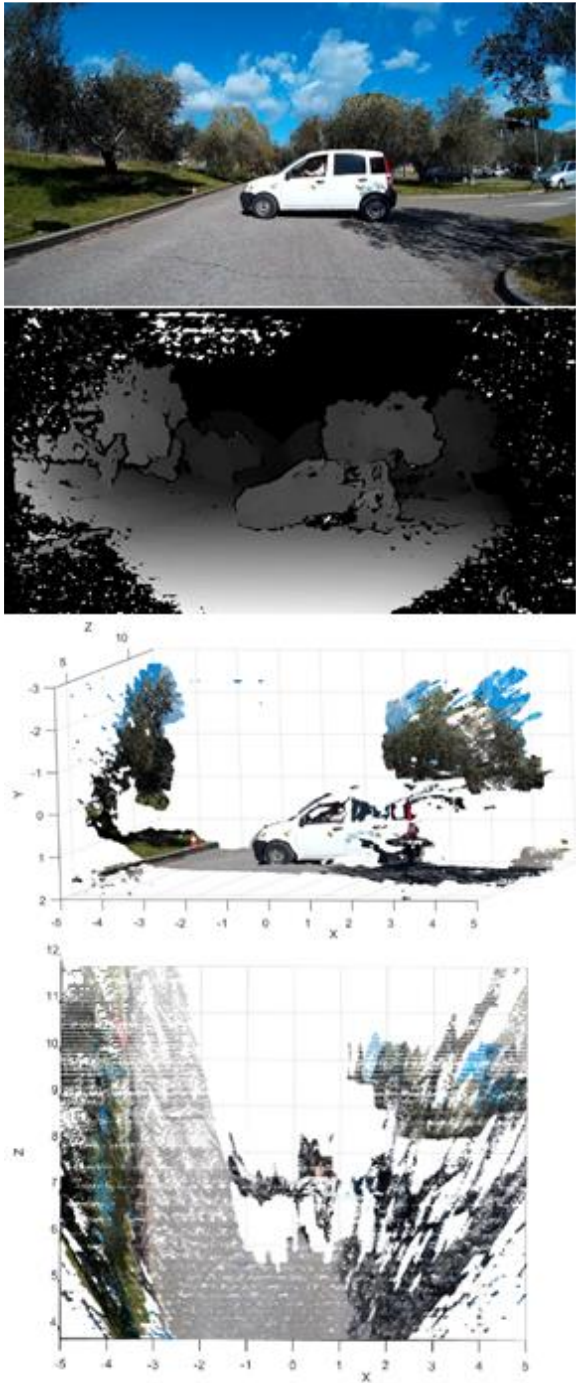

*Figure 7. Measurement of the heading angle of the Opposite Vehicle (OV). In this sequence, our test car include 5 visible detection markers (10 cm diameter).*

We place detection markers along the surface of our test vehicle. The detection markers are used as landmarks in the car itself (Figure 7, and more in Figures 11 and 12), allowing to use the same video frames of the cameras to conduct photogrammetric analysis and compare the heading angle measured from the stereo vision system and the laser scanner.

The following three results show part of the calibration activity with the aim of refining the measurement of the heading angle.

**Stationary field testing**
This experiments were conducted with a stationary setup as used in the previos cases, but in the real traffic (Figures 13-15). As in this case much less information about the ground truth of the scene is known, we decided to analyze an urban roundabout in which the range of speed and trajectories are more homegeneus than in a normal intersection, contributing to our scope of sensing the heading of opposite vehicles.
Figure 14 shows the detection of a second vehicle (a bus). As the bus is out of the range of measurement of this central baseline, only the frontal part of it was reconstructed. However the gap car-bus was well measured. In Figure 15, the whole bus is in the measurement range therefore we could measure the length of the vehicle (12 m) with our stereo system.

**Moving field measurements**
The following results refer to moving cameras mounted on our test PTW. The indesireable noisy effect in all the disparities maps concerning to the asfalt became more noticeable. This effect is produced by the motion blur of the cameras (see Figure 16). Further activity is required in the quantification of the vibrations at which the cameras are subjected because this effect is remarkable on PTWs in comparision with cars.
In the following case (see Figure 17) we employed the well-defined box of the lorry to assess the heading and the measurement of a large planar surfaces.
In the last measurement conducted with moving PTW (Figure 18), we verify that narrow road users can be properly detected with the stereo vision system. In the disparity map it can be seen that both PTWs in the scene are well measured; the lorry appears in the edge of the range of measurement; the ground surface appears noisy. The scooter on the left appears very well defined. This is because the relative velocity with the host PTW is almost zero and from the point of view of the imaging system it is similar to a static object. On the contrary, the PTW on the right side is parked in the curb and the relative velocity corresponds to the velocity of the host PTW (40 km/h). Notwithstanding, it is also measured well as it can be seen from the disparity map and from two perspectives of the point cloud reconstruction.

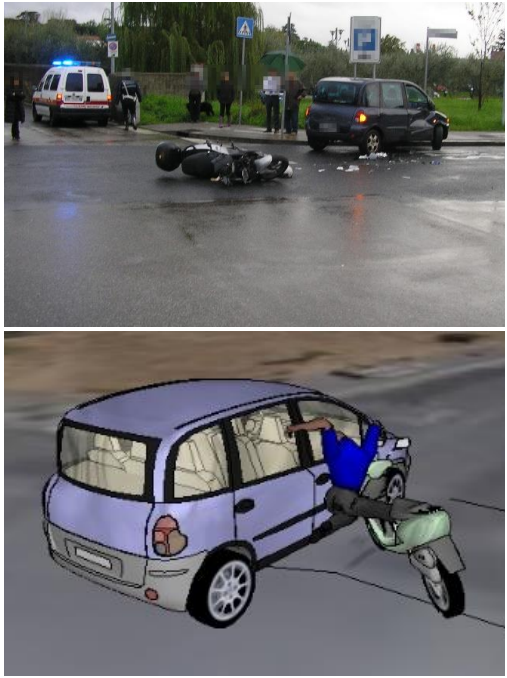

*Figure 9. PTW and opponent vehicle final positions (top) and impact configuration (bottom).*

**Test case description (crash ID86)**
With the given selection criteria 11 PTW crash cases were identified from InSAFE. The present paper focused on one case, the InSAFE ID86 (Figure 9). The crash occurred on a rainy and cloudy afternoon. The opponent vehicle was a FIAT Multipla and the driver was approaching a crossing, without priority, coming from the left side of the PTW. The Aprilia Sportcity rider, with priority on the driver, went straight at the crossing. The rider and the driver were travelling at 55 km/h and 30 km/h, respectively. No mobile or fixed obstacles obstructed the drivers' field of view. The driver did not halt at the stop sign and passed through. Around 1 second before the point of impact, the rider took a pre-impact avoiding action, thus slowing down up to 45 km/h with an estimated acceleration of -2.8 m/s$^2$, applied 1 s before the impact. The PTW collided frontally with the right side of the opponent vehicle. The rider was wearing an open face helmet during the crash and suffered head (MAIS2) and spinal injuries (MAIS2).For this particular case, the activation of the Emergency Braking would have occurred 7.2 m to the collision point. The case was analyzed considering the PTW located at a distance of 8 m from the point of collision.

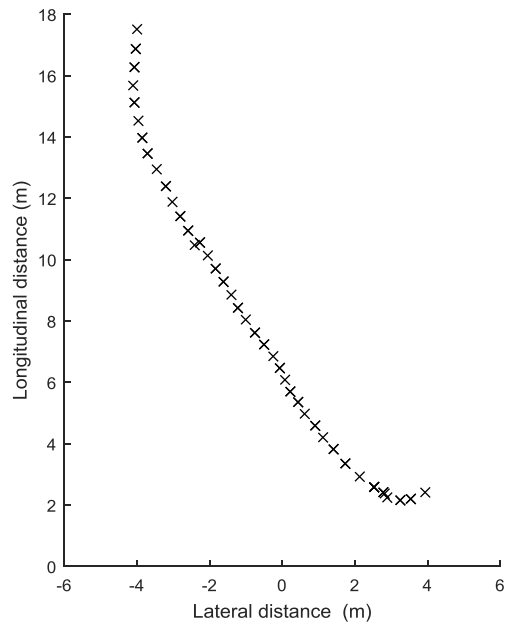

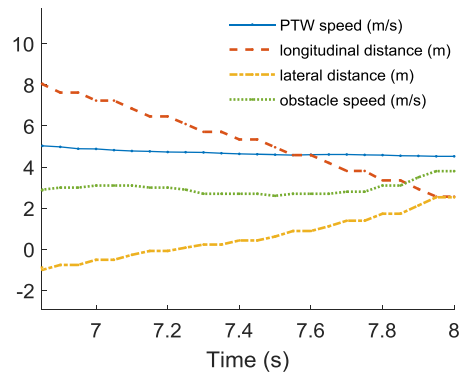

*Figure 8. Relative position coordinates (top) and other state parameters (bottom) of the opponent vehicle detected from the laserscanner - case ID86.*

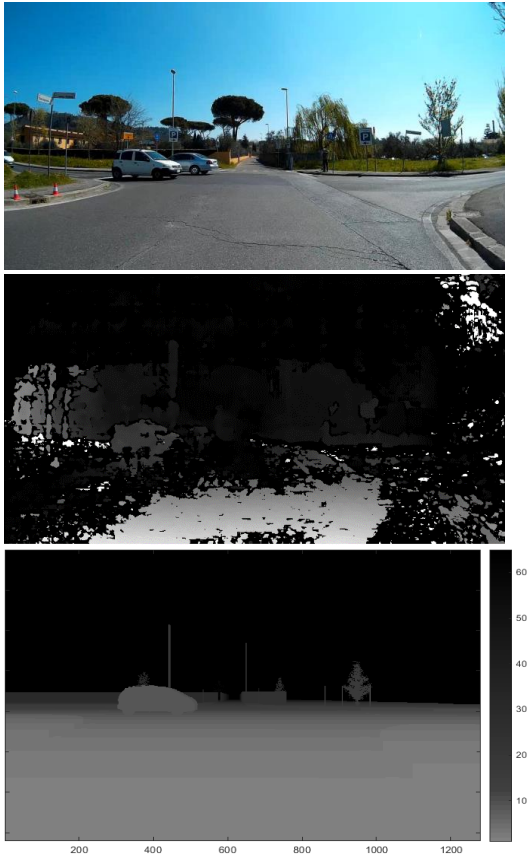

*Figure 10. Emulation of crash case ID86 (top figure). The disparity map acquired during the emulation (middle) is compared with the instrumented vehicles and the synthetic ground truth simulated with PreScan (bottom).*

The laserscanner was able to detect the position coordinates of the opponent vehicle (see Figure 8). However, the heading angle of the opponent vehicle was not correctly measured by the laserscanner, which produced erratic oscillations between 10 degrees and 90 degrees (not shown in the figure).

The results depicted in Figure 10 focus on the stereo analysis of the wider field of view of the artificial vision system. From top to bottom, we can see one rectified image of the scene, its disparity map and a detail of the depth measured from the virtual environment. The disparity map computed from the stereo cameras is noisy, due to the motion of the host vehicle. However it is possible to identify an homogeneous volume corresponding to the lateral part of the opponent vehicle, from which heading angle can be estimated.

## CONCLUSIONS

The encouraging preliminary results of the stereo vision approach suggest that such application of stereo vision is suitable to address this kind of common PTW crashes at intersections. The tremendous evolution of camera sensors present in mobile phone and portable devices industry makes stereo vision technology attractive for the motorcycle field. In fact, even if cameras cannot measure objects through fog or rain, PTW crashes often occur in good visibility conditions.
Several remarks about the degradation of the disparity map were pointed out during the presentation of the results. Further activities to address these issues are warranted to improve real world applicability.

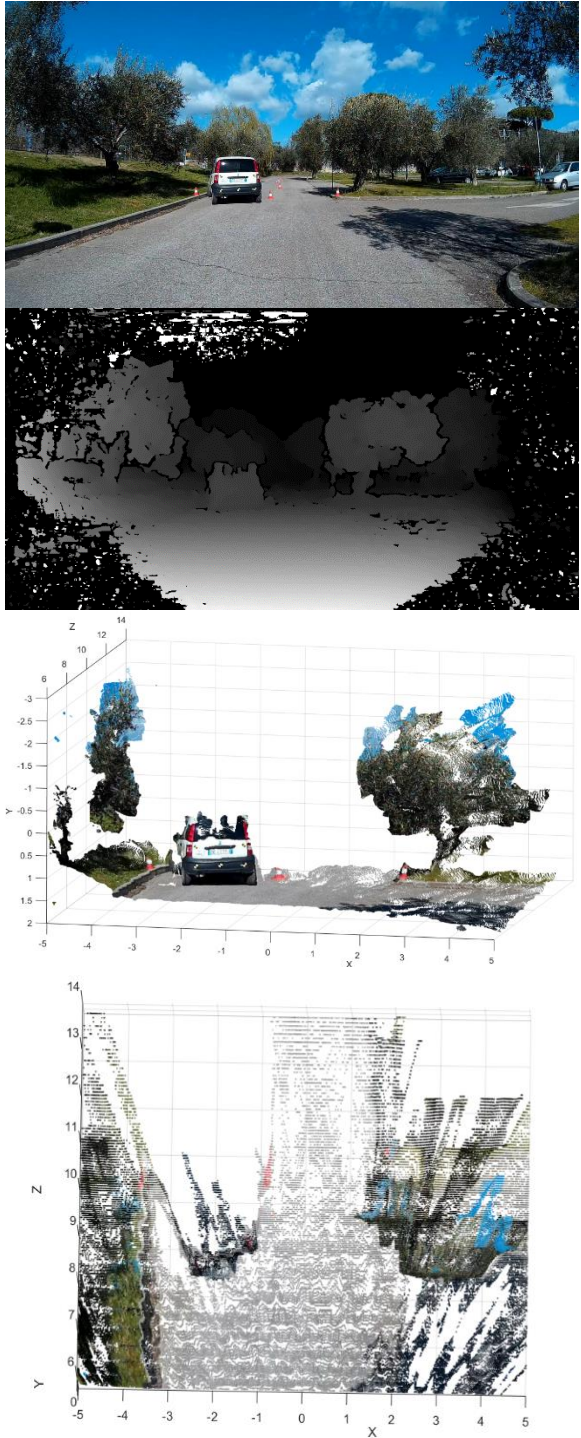

*Figure 11. Measurement of the heading angle of the Opposite Vehicle (OV). In this sequence, our test car include 5 visible detection markers (10cm diameter).*

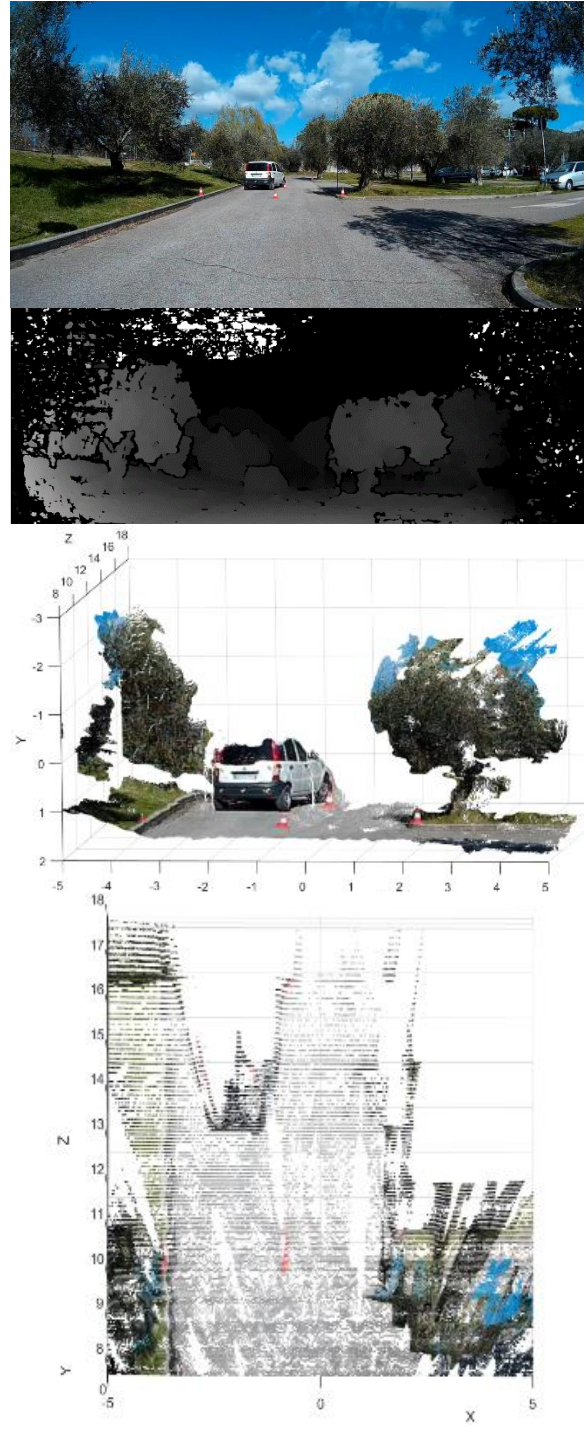

*Figure 12. Measurement of the heading of the OV.*

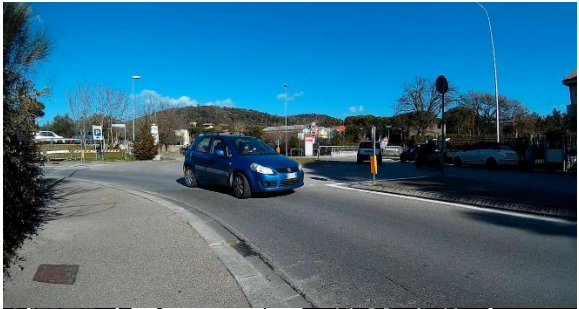
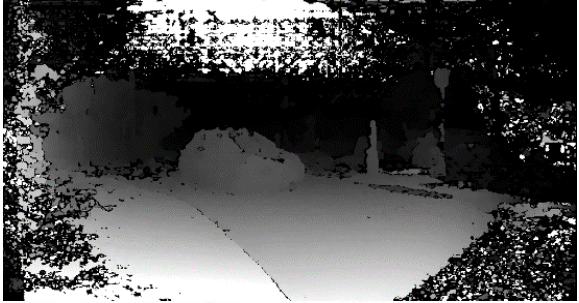
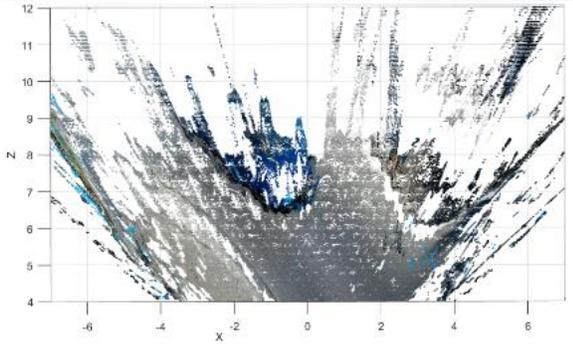
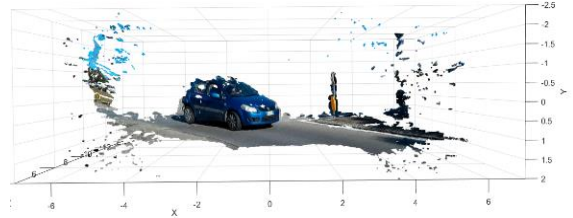

*Figure 13. Measurement of the heading of the OV.*

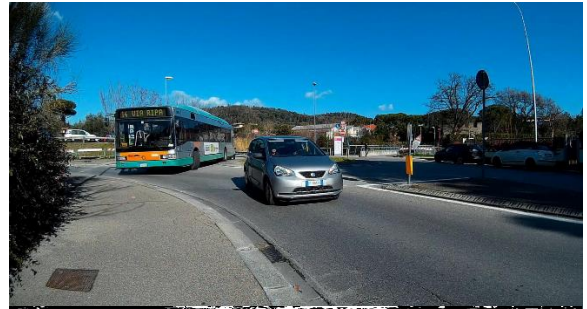
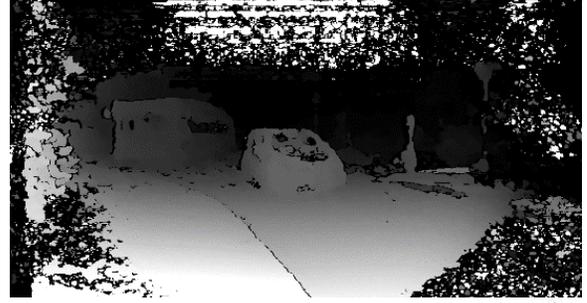
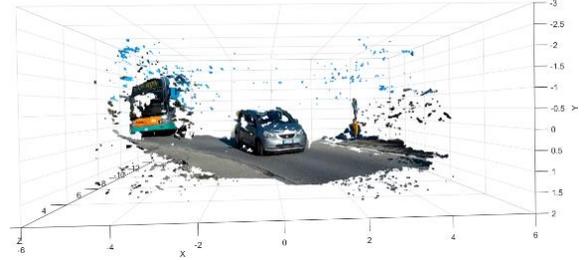
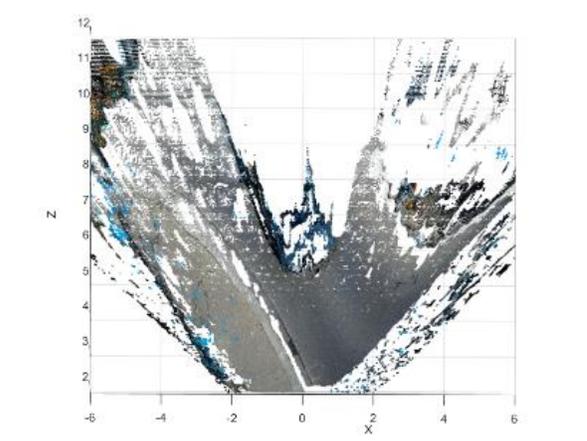

*Figure 14. Measurement of two vehicles.*

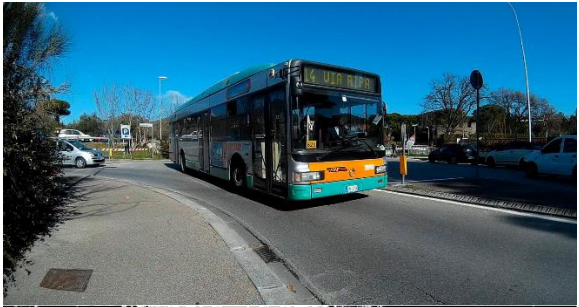
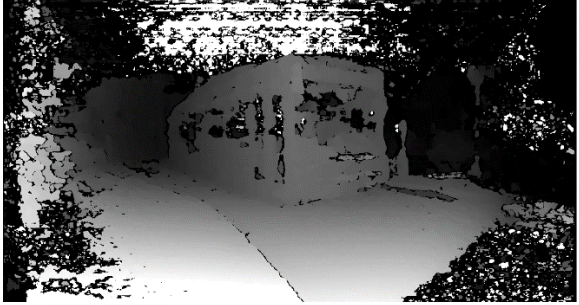
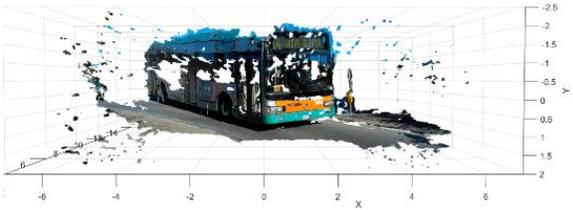
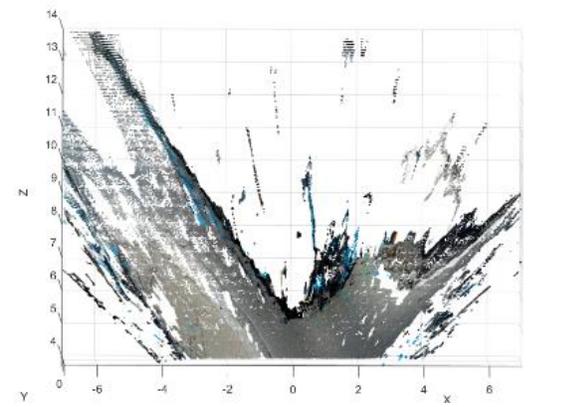

*Figure 15. Heading of big obstacles.*

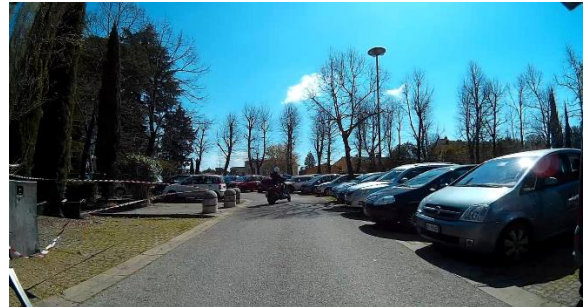
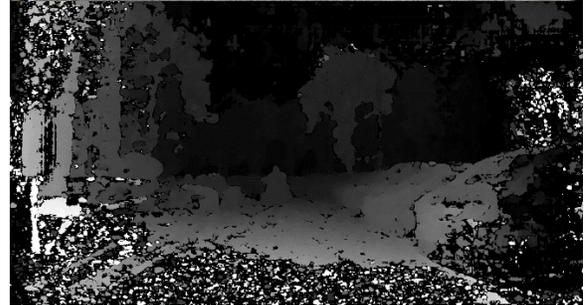
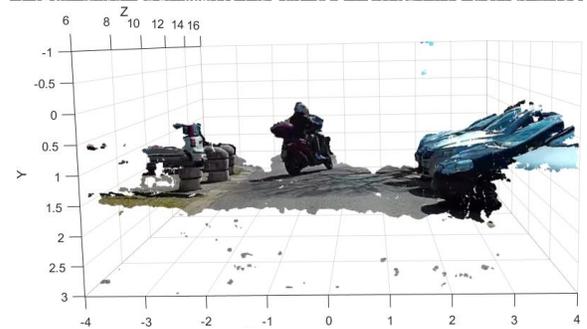
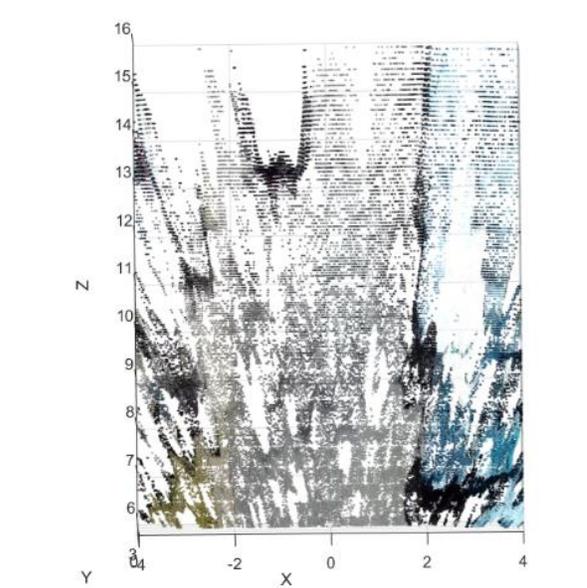

*Figure 16. Measurement of other PTW entering traffic.*

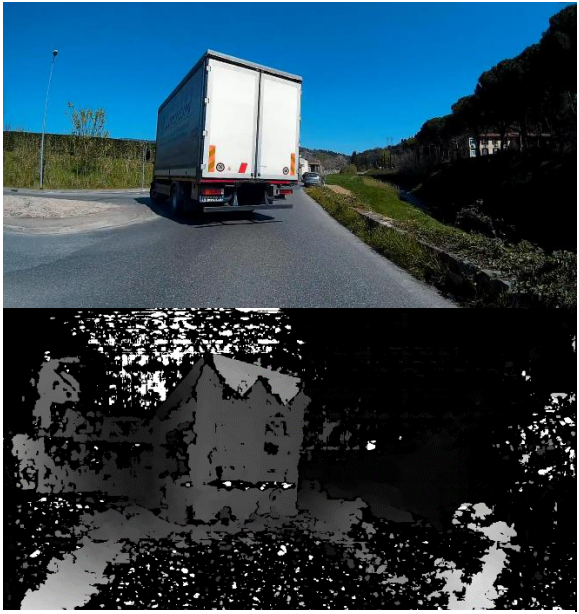

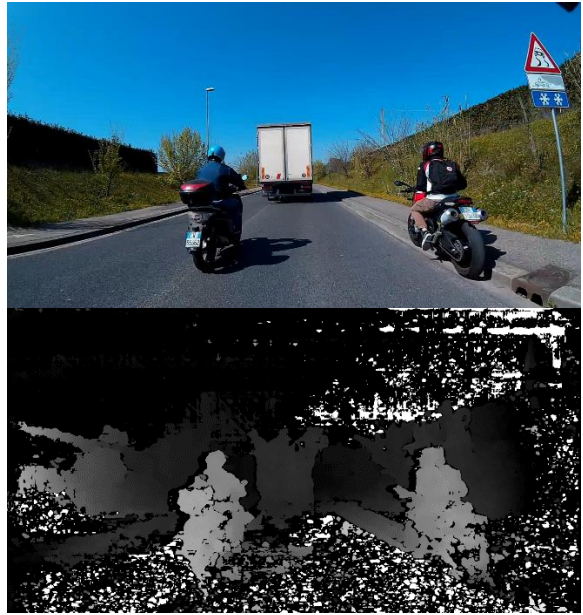

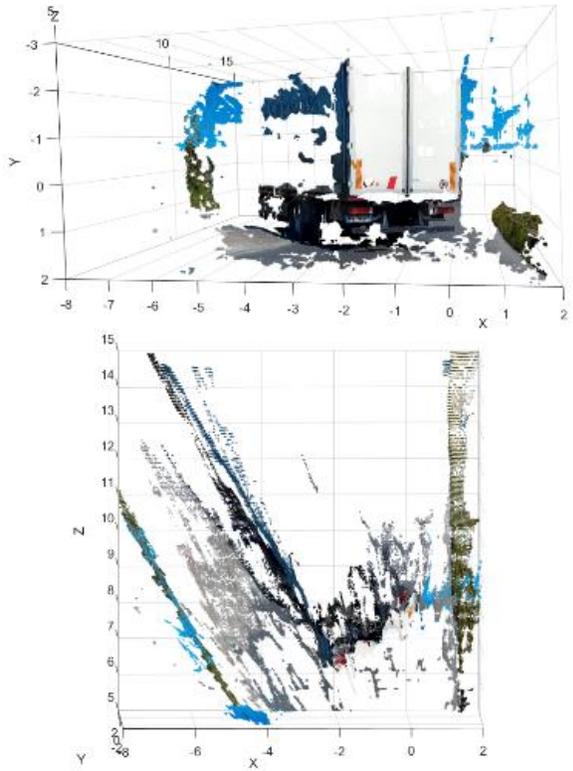

*Figure 17. Measurement of a large obstacle.*

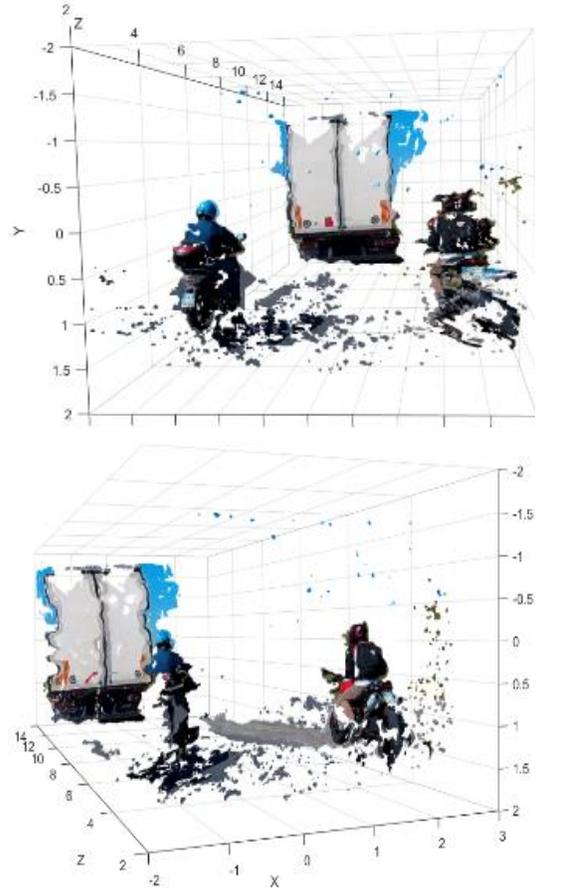

*Figure 18. Measurement of two narrow objects.*